\def\year{2020}\relax
\documentclass[letterpaper]{article}
\usepackage{aaai}
\usepackage{times}
\usepackage{helvet}
\usepackage{courier}
\usepackage{graphicx}
\usepackage{amsmath}
\usepackage{amsfonts}
\usepackage{subfigure}
\usepackage{color}

\graphicspath{figures/}

\begin{document}
%
\title{Model-based Reinforcement Learning for Predictions and Control for Limit Order Books}

\author{Haoran Wei,\textsuperscript{1*}
Yuanbo Wang,\textsuperscript{2*}, Lidia Mangu,\textsuperscript{2**}, Keith Decker,\textsuperscript{1**}\\
\textsuperscript{1}{University of Delaware}\\
\textsuperscript{2}{J.P. Morgan}\\
nancywhr@udel.edu,
yuanbo.wang@jpmchase.com, 
lidia.l.mangu@jpmchase.com
decker@udel.edu}
\maketitle
\begin{abstract}
\begin{quote}

We build a profitable electronic trading agent with Reinforcement Learning that places buy and sell orders in the stock market. An environment model is built only with historical observational data, and the RL agent learns the trading policy by interacting with the environment model instead of with the real-market to minimize the risk and potential monetary loss. Trained in unsupervised and self-supervised fashion, our environment model learned a temporal and causal representation of the market in latent space through deep neural networks. We demonstrate that the trading policy trained entirely within the environment model can be transferred back into the real market and maintain its profitability. We believe that this environment model can serve as a robust simulator that predicts market movement as well as trade impact for further studies.  


\end{quote}
\end{abstract}

\section{Introduction}

Model-free Reinforcement Learning (MFRL) using deep-learning architecture has achieved robust performance on a variety of complicated tasks, ranging from the classic Atari 2600 video games \cite{mnih2015human}, to locomotion tasks \cite{lillicrap2015continuous}, and have even delivered super-human performance in challenging exploration domains \cite{salimans2018learning}. Learning by trial and error, one of the limitations of such approaches is that they usually require a large number of interactions with the environment for training purposes. This challenge becomes even more significant when these interactions are expensive or even dangerous (e.g., financial trading or self-driving cars). 

Building a simulator of the environment could alleviate such problems. By learning a dynamic model that predicts the next state given current state and action, model-based RL (MBRL) enables agents to explore inside the simulator safely. In recent research, methods have proved to be sample efficient in various tasks without compromising policy effectiveness compared to model-free approaches \cite{kaiser2019model,zhang2018solar}. Additionally, an environment model that enables predictions of the future is not only appealing to RL but also has general applications across various domains.

Most of the MBRL work still relies on using some interactive feedback from reality to update the simulator online. This is crucial to avoid mismatching problems: the approximation error of the environment model may lead the RL policy in the wrong direction. With a fixed data set to train policy with, this deviation from optimal strategy becomes challenging to self-correct. Recent studies have started challenging this requirement for real-time interactions with the environment. David Ha et al\cite{ha2018world} experimented with training policies for the VizDoom game completely inside the environment model and it outperformed benchmarks after being transferred back into the original game. Similarly, robust performance was observed in applying model-based RL methods to learn car-driving policies using observational data \cite{henaff2019model}. To advance MBRL systems without real environment interactive data in the more complex real world, we propose a model-based RL framework and test its robustness in electronic trading domains.  

Interestingly enough, we observe similarities between the implicit structures of MBRL and the explicit design of an electronic trading system. In the most general and simplistic case, an electronic trading system consists of a market simulator and decision strategies. The simulator predicts market dynamics using quantitative and statistical models, and it has inspired extensive research\cite{ang2006stock,bacchetta2009predictability}. The decision strategies are usually influenced by domain knowledge, heuristics, observations, and sometimes, preferences of the algorithm users. There are two main types of approach to building market simulators: traditional statistical methods and machine learning methods. Traditional statistical approaches try to model linear processes that underlie the generation of time-series. For instance, the Glosten model \cite{rocsu2009dynamic} assumes that the price impact of trades is linear, immediate, and not state-dependent. The Vector Auto-Regressive model (VAR) \cite{zivot2006vector} addresses the non-stationary aspect of financial time-series and can forecast with multiple variances. Similarly, in \cite{beltran2005understanding}, researchers investigate the near-linear dynamics between sequences of orders and the evolution of the market. On the other hand, machine learning approaches, which do not assume linearity and require little prior knowledge about the input data, have also shown promising performance in market forecasting \cite{zhang2019deeplob,ntakaris2018benchmark,tsantekidis2017forecasting,tsantekidis2017using,dixon2017classification}. 

Two drawbacks in this explicit simulator-policy framework are a compilation of approximation errors from the simulator and explosion in the number of corner cases in hand-crafted decision policies. There have been efforts to bypass the market-simulating stage and proceed to train an RL agent that takes market conditions as input and directly outputs decisions. \cite{bacoyannis2018idiosyncrasies} applies a model-free technique to solve trade execution problems. \cite{nevmyvaka2006reinforcement} trains an RL policy to replace hand-crafted decision rules while still employing a traditional market simulator. However, the same challenge persists that RL agents are still trained using static, pre-generated time-series data.

Given the structural similarities between MBRL and electronic trading systems, we think a model-based RL agent trained on observational data could potentially solve the problems in electronic trading that we have listed. To our knowledge, no prior work has been done on this topic, and therefore, we decide to bridge the gap. We have two main contributions: (1) In our MBRL framework, we use latent representation learning to model not only the state space but also rewards. We demonstrate the effectiveness of such representation learning in the financial domain, where data is high-dimensional and non-stationary. (2) Using this model of the environment, we show that our model-based agent consistently outperforms commonly used benchmark trading strategies. This approach enables the learning of profitable trading policies using observational data with \textit{no} environment interaction or labeling by human experts. This project code will be released soon for replication.

\section{Related Work}

Two studies inspired our study. In World Models \cite{ha2018world}, the authors employ generative models to get latent representations of the environment, which enables the RL agent to learn a compact yet effective policy. The other recent study builds a neural network simulated environment $\text{env}_{0}$ that not only shares an action space and reward space with the original environment $\text{env}$ but also produces observations in the same format \cite{kaiser2019model}. 

In reality, there are many scenarios where real environment interactions are costly or not feasible, such as autonomous driving \cite{wu2017flow}, recommendation systems\cite{zhao2019model}, and trading systems. These infeasible interaction environments yield a challenge for RL. However, if the next state can be predicted, real environment interactions may not be necessary anymore. Lukasz Kaiser et al. \cite{kaiser2019model} show a complete model-based RL approach to play Atari games where a CNN-based prediction model is used to predict the next game frame given the previous frames and action, and CNN layers are used to extract hidden features autonomously. A variational layer is used as the last layer to learn the posterior of the next frame context so that environment stochasticity is considered, and learning is shown to be improved\cite{ha2018world,oh2015action,leibfried2016deep}. Recently, model-based RL is also used in recommendation systems \cite{zhao2019model}  for conducting random exploration without bothering users overwhelmingly. However, few works have been seen in real-world applications compared to the wide application in the gaming domain. This is reasonable because the real world has more complicated, uncertain factors to model. 

In the Finance domain, RL has been applied to many different problems \cite{fischer2018reinforcement}, especially designing electronic trading strategies \cite{bacoyannis2018idiosyncrasies,bertoluzzo2012testing}. However, most of the work has been done with model-free RL, such as Deep Q-networks (DQN)\cite{huang2018financial}, that have lower sampling complexity. Alternatively, model-based methods require many fewer training samples; however, there is no existing finance RL model for random exploration. Our work tries to show that 1) A trading model can be built with historical observation data; 2) model-based RL has potential in time-series decision making.

\section{Background}
In this section, we introduce some important elements in electronic trading systems and how they translate to concepts familiar to RL research. We also give a brief overview of the RL methods used in this study. 

\subsection{Trading Problem}

\begin{figure}
    \centering
    \includegraphics[width=0.3\textwidth]{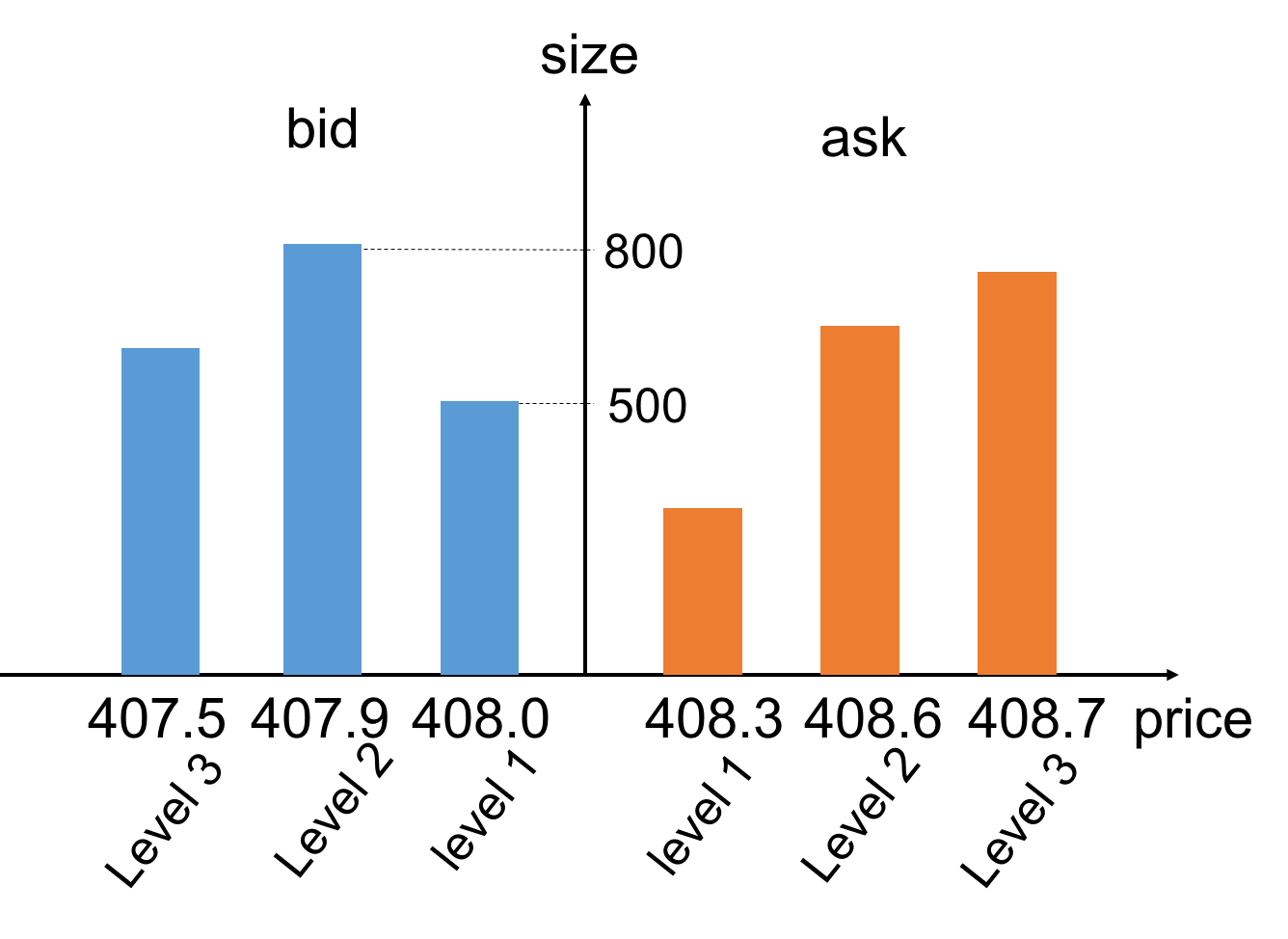}
    \caption{A slice of limit order books (LOB) with three levels on both ask and bid sides}
    \label{lob_demo}
\end{figure}

\textbf{Limit Order Books} (LOBs) are used by more than half of the exchange markets in the world \cite{rosu2010liquidity}. An LOB has two types of orders: bid and ask. At any given time $t$, a bid (ask) order is to buy (sell) certain quantity (aka size), $\text{bs}(t)$ ($\text{as}(t)$), of a financial instrument at or below (above) the specifies price $\text{bp}(t)$ ($\text{ap}(t)$), as shown in Fig \ref{lob_demo}. Orders submitted at time $t$ are sorted into different levels based on their prices. For instance, the lowest ask price and the highest bid price are grouped into the first level order, followed by the second lowest \text{ap} and the second-highest \text{bp} as the second level, and so on. $\left\{ \text{ap}, \text{bp}, \text{as}, \text{bs}\right\}_{i}$ are vectors of values and quantities at different price levels $i$. The time-series evolution of an LOB can be seen as a 3-dimensional tensor: the first dimension represents time, the second dimension is level, and the third represents prices and order quantities on both the buy and sell sides\cite{gould2013limit}. When submitted orders are executed by an LOB's trade-matching algorithm, the orders' price and quantity with direction (bid or ask) are removed from the LOB and recorded in a historical \textit{\textbf{trade print}}. 

\textbf{Mid Price}  is the mean value of the first-level ask and bid price, as Eq \ref{midprice}
\begin{equation}
    \text{mid}_{t} = \frac{\text{ap}_{1} + \text{bp}_{1}}{2}
\label{midprice}
\end{equation}
Take the slice of LOB in Fig \ref{lob_demo} as an example, the mid price at this time $t$ is (408.0+408.3)/2 = 408.15. The movement of the mid price is commonly used to approximate market change. In this study, we use the mid price to calculate reward.

\textbf{Trade prints} are the record of executed trades that contain information of direction (buy or sell), trading price, and quantity. The collection of trade prints may be executed by different agents in the market. In this study, we use historical trade prints as our RL agent's exploration actions. We also include a sequence of trade prints prior to the target action as part of the state. This data provides crucial information for the state transition probabilities model (details are in the next section). We observe a similar problem setup in \cite{henaff2019model}, where the authors use a target car's driving trajectory for RL exploration and surrounding cars' trajectories as part of states.

\subsection{RL for agent}
We test three commonly used RL algorithms for policy training: Double Deep-Q network (Double DQN) \cite{van2016deep}, Policy Gradient \cite{williams1992simple}, and Advantage Actor-Critic (A2C) \cite{mnih2016asynchronous}. 

\textbf{Double DQN} 
A deep Q-network is a neural network that approximates the Q-value for an input state-action pair. It's optimized by minimizing the square error between the predicted and target Q-value. Double DQN uses two networks: an online network for selecting the actions according to the value of ($Q, \theta$), and another target network to determine its value ($Q, \theta_{-}$). Optimization of the online network ($Q, \theta$) is done through minimizing the prediction error $L(\theta)$ of the target Q-network:
\begin{equation}
\begin{aligned}
   & L(\theta) = \mathbb{E}[(Q(s, a;\theta) - Q_{\text{target}})^{2}]\\
   & Q_{\text{target}} = r+\gamma Q(s', \arg\max_{a'}Q(s', a'; \theta); \theta_{-})
\end{aligned}
\end{equation}

$Q_{\text{target}}$ is the output of target network ($Q, \theta_{-}$), the weights ($\theta_{-}$) of which remain unchanged except for a periodic copy of weights from the online Q-network ($Q, \theta$). Having a separate target Q-network helps reduce policy variance caused by oscillations of the target value.

\textbf{Policy Gradient} In the policy gradient (PG) algorithm, a policy is directly modeled with a $\theta$-parameterized function $\pi(a|s,\theta)$. Given $\pi(a|s,\theta)$ and environment model $p(s'|s, a)$, we can generate a trajectory $\tau=(s_1, a_1, s_2, a_2, ... s_t, a_t)$ and accumulated reward $r$, see Eq\ref{PG}. The idea is to maximize the accumulated reward $J(\theta)$ by repeatedly optimize $\theta$ with gradient ascent $\nabla J(\theta)$. 
\begin{equation}
\begin{aligned}
     &J(\theta) = \mathbb{E}_{\tau}[\sum_{t=0}^{H}\log\pi(a|s,\theta)G_{\tau}]\\
     & \text{where} \qquad G_{\tau} = \sum_{t=0}^{H}\gamma^{t}r_{t}
\end{aligned}
\label{PG}
\end{equation}

Compared to other value-based RL methods, PG learns a policy directly. It is also less sensitive to value overestimation problems common in Q-learning (caused by the ``max" operation). However, one drawback is that reward accumulation along a trajectory may cause high policy variance. 

\textbf{Advantage Actor-Critic (A2C)} A2C is a hybrid RL method combining policy gradient and value-based methods. It consists of two networks: actor and critic. The actor network updates policy $\pi(a|s,\theta)$ by maximizing the objective function:
\begin{equation}
J(\theta) = \mathbb{E}_{\tau}[\sum_{t=0}^{H}\log\pi(a|s,\theta)A(s, a)]
\end{equation}
where $A(s, a)$ is the advantage value representing how much better (worse) a given action performs in state $s$ compared to the average performance over all actions. It updates as:
\begin{equation}
A(s_{t}, a_{t}) = r_{t} + \gamma V(s_{t+1};w) - V(s_{t};w)
\end{equation}
$V(s;w)$ is the state utility representing the average performance over all actions in a given state, computed by the critic network. The critic network's parameter $w$ is updated by gradient descent on the TD-error (parameterized with $w$):
\begin{equation}
J(w) = (r_{t} + \gamma V(s_{t+1};w) - V(s_{t};w_{-}))^{2}
\end{equation}
$w_{-}$ is the parameter from the critic's previous update. The advantage of A2C is twofold: 1) policy variance is reduced due to the advantage value; 2) the policy is directly updated instead of via a value estimation function. 

\section{Problem Formulation and Dataset}
\label{Problem Formulation and Data Preparation}
Limit Order Book data is time-series with high sampling frequency. We model the environment as a Markov Decision Process (MDP). We use time-series data (LOB + trade prints) from one stock traded on the Hong Kong Stock Exchange. The update frequency of the LOB is $\sim$ 0.17s. We use January 2018 $\sim$ March 2018 (61 days) data for training with 20\% as the validation set, and test model performance on April 2018 (19 days). In all, the dataset contains approximately 6 million transitions for training and 2 million for testing.

\subsection{MDP Forming}
We model the trading problem as a Markov Decision Process (MDP) represented as $\{\mathcal{S}, \mathcal{A}, \mathcal{R}, \mathcal{T}, \rho_{0}\}$. We'll explain how to build the MDP with trading elements in this section. 

\textbf{State Space ($\mathcal{S}$)}: $s_{t} = \left\{\text{ae}(\text{ob}_{t-T:t}), u_{t},\text{po}_{t}\right\}$. $z_{t}=\text{ae}(\text{ob}_{t-T:t})$ is a latent representation of the LOB within a time duration $T$. We will discuss the latent representation model more in the following section. $u_{t}$ is a vector of trade prints occurring within the same duration $T$.  $\text{po}_{t}$ is the RL agent's position at time $\text{t}$. Position reflects the private inventory held by the agent, which, in our case, is bounded by $(-\text{po}_{\text{max}}, \text{po}_{\text{max}})$. 

\textbf{Action Space ($\mathcal{A}$)}: $a_{t} = \pm q$. Each action is the RL agent's decision to trade. The decision includes price, quantity and direction of the trade. In our study, we assume that the trading price is set at mid-price, and can be directly calculated from the LOB update. Therefore the RL agent's action contains $q$ the absolute value of quantity and $\pm$ trading direction (sell/buy). 

\textbf{Reward Function ($\mathcal{R}$)}: We use a mark-to-market PnL to calculate agent's reward. It's defined as:
\begin{equation}
    \mathcal{R}(t) = \Delta\text{mid}_{s_{t}, s_{t+1}} \times \text{po}_{t}
\label{reward_func}
\end{equation}

where $\text{po}_{t}$ is the RL agent's position at time $t$ and $\Delta\text{mid}_{s_{t}, s_{t+1}}$ is the difference in average LOB mid-price between state $s_{t}$ and $s_{t+1}$. 

\begin{figure}
 \centering
\includegraphics[width=0.35\textwidth]{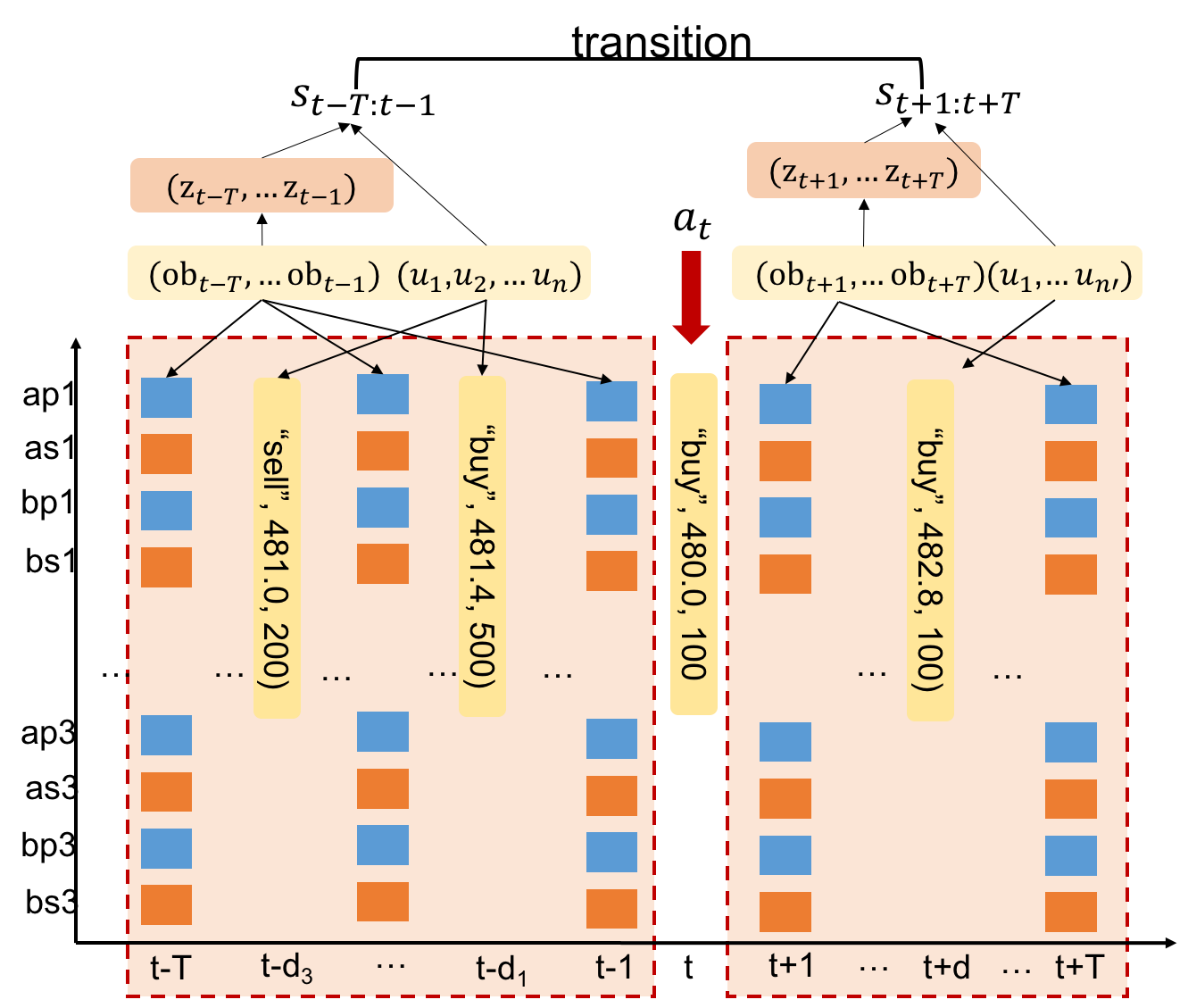}
\caption{A demonstration of forming a state-action-state transition with limit order books and trade prints (the subscriptions are the time ticks for the lob and are indices only for the trade prints)}
\label{transition_built}
\end{figure}

\textbf{Transitions ($\mathcal{T}(s_{t+1}|s_{t}, a_{t})$)}. We use observed trajectories of state transitions to train the environment model, and one transition is demonstrated in Fig \ref{transition_built}. Specifically, we iterate through trade prints in historical data and treat each trade as a target action in the transition. The same action may become a part of the state $u_{t}$ in the next transition when the next trade becomes the target action. 

\textbf{Initial State} ($\rho_{0}$): The initial states are sampled from the first state over all days in the training dataset following a uniform distribution. 

\subsection{Data Preprocessing}

For each LOB time-series sequence $\text{ob}_{t-T:t}$ with a length of $T$, we use the feature-level min and max to normalize the data across time. For the trade quantity normalization, we first exclude the outlier trades that either has less than 100 or exceed 1000 of quantity. Then we use 100 and 1000 as the boundary for min-max normalization, and the boundary is empirically determined based on the data distribution. We also implement the min-max normalization followed by a sigmoid transformation on the rewards. Here it's worth noticing that the reward transformation is only done inside the world model during exploration for training purposes. 

\section{Model}
This work has two parts: a world model that consists of latent representation learning of the LOB, state-action transition, and reward models. A trading agent trained based on the world model with three widely used RL methods. 

\begin{figure*}[t]
\centering
\subfigure[]{\label{ave}\includegraphics[width=0.3\textwidth]{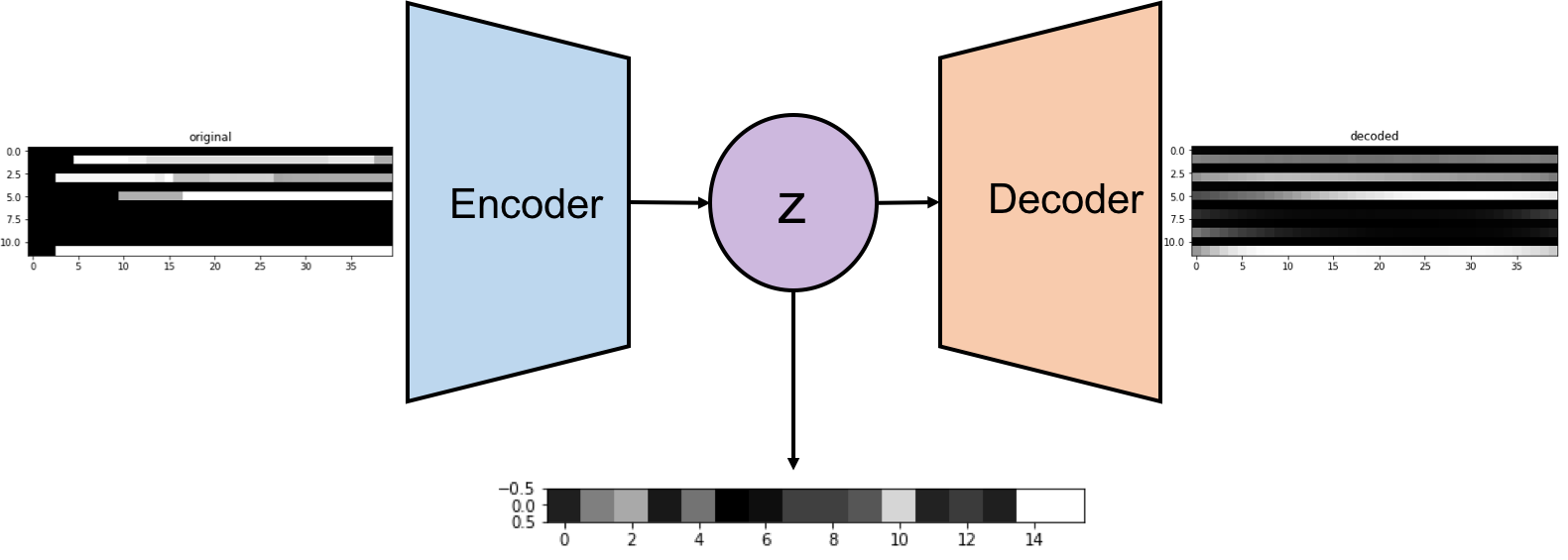}}
\subfigure[]{\label{rnn_mdn}\includegraphics[width=0.3\textwidth]{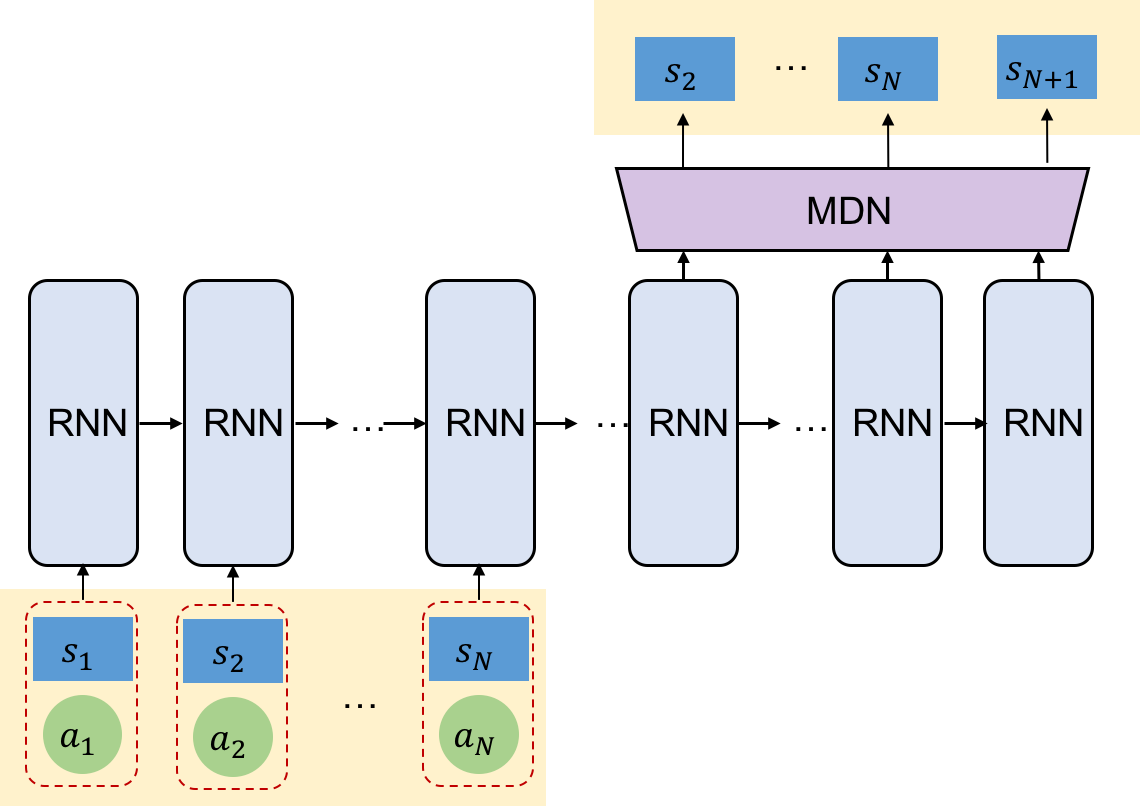}}
\subfigure[]{\label{workflow}\includegraphics[width=0.3\textwidth]{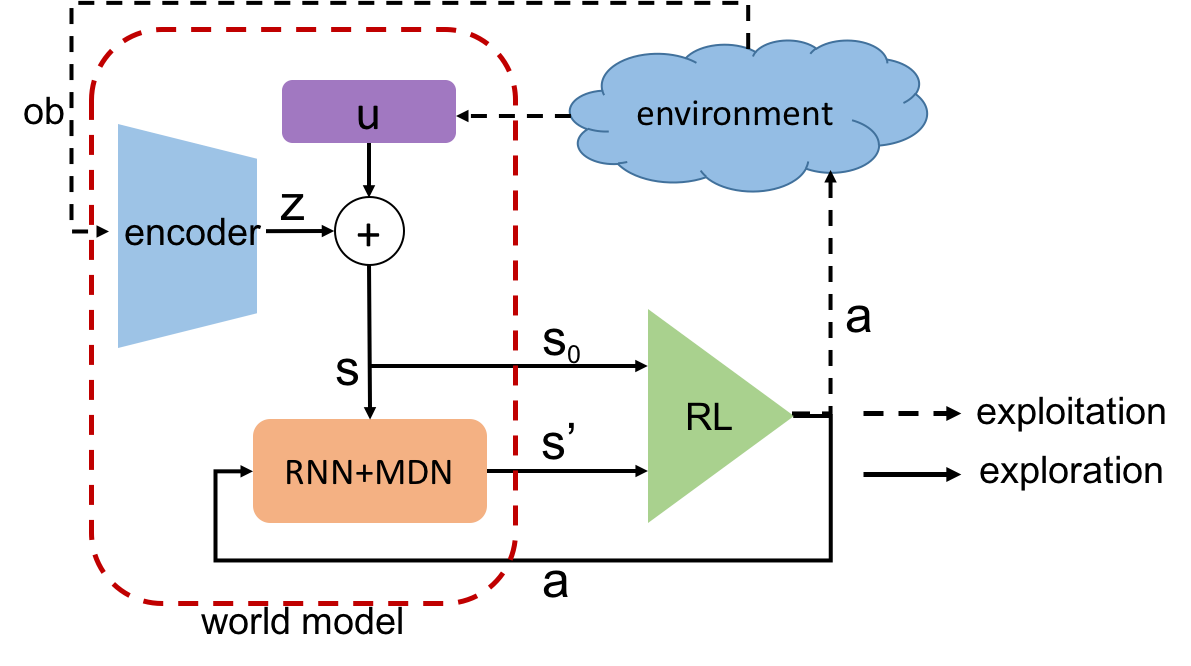}}
\caption{(a) Extract latent LOB representation with an auto-encoder (AE) (b)The transition model: a sequence-to-sequence RNN-MDN (c)The RL agent's workflow: exploring with the environment model and exploiting in the real environment.(The dashed lines represent the exploitation and solid lines represent the exploration)}
\end{figure*}

\subsection{World Model}
The world model consists of 3 parts: a latent representation model, transition model, and reward model.

\subsubsection{Latent Representation Model (Auto-encoder)}
The LOB data contains important time-series market information but it's difficult to learn due to its high dimensionality. The role of the Auto-Encoder (AE) (as shown in Fig \ref{ave}) is to find an abstract and low-dimensional representation of the LOB observations for easy learning. The input and output of AE are both of dimension $T \times 4L$ where $L$ is the number of levels in LOB. We use 3 levels in our experiments. After AE is trained, we take the middle layer of dimension $1 \times m$ as the latent representation of high-dimensional LOB data. Here we have $m=16$. 

\subsubsection{Transition Model (RNN-MDN)}
We use a RNN-MDN model to learn the state-action transitions $\mathcal{P}(s'|s, a)$. This model also addresses trades' impact on the market. We use a long sequence of \text{s, a} to train RNN and learn both short term and long term impacts. We further combine the RNN with a Mixture Density Network (MDN) \cite{bishop1994mixture}. This approach approximates the output of RNN as a mixture of distributions rather than a deterministic prediction of $s'$. The MAD is written as:
\begin{equation}
    p(s'|s, a) = \sum^{K}_{k=1}w_{k}(s, a)\mathcal{D}(s'|\mu_{k}(s, a), \sigma^{2}_{k}(s, a))
\end{equation}
where $\mathcal{D}(\cdot)$ is a presumed distribution, such as Gaussian, Bernoulli, etc. The length of input sequence to the RNN is $N$, and the $(N+1)^{\text{th}}$ state is the prediction based on the previous $N$ sequences, shown as Fig \ref{rnn_mdn}. This approach has been widely applied in the past on sequence-generating tasks \cite{ha2017neural}.

\subsubsection{Reward Model}
Since the PnL depends on the next state that is unseen for the RL agent, a regression model is used to predict the change of mid-price based on the current latent state and the predicted next latent state:
\begin{equation}
    r_{t} = \mathcal{R}(z_{t}, z_{t+1};\beta) \times \text{po}_{t}
\end{equation}
where $\beta$ is the reward model's parameter. 

In the reward model, position $\text{po}$ depends on the executed actions and a hand-crafted capacity that casts the limit. The next position given current position and action is calculated as:
\begin{equation}
    \text{po}_{t+1} = \left\{\begin{matrix}
 \min(\text{po}_{t}+\left | a_{t} \right |, \text{po}_{\text{max}})& \text{when} & a_{t}>0\\ 
 \max(\text{po}_{t}-\left | a_{t} \right |, -\text{po}_{\text{max}})& \text{when} & a_{t}<0 
\end{matrix}\right.
\end{equation}
where $(-\text{po}_{\text{max}}, \text{po}_{\text{max}})$ defines the position capacity. 

The latent representation, transition model, and reward model together are the world model. Then, we train the RL trading agent purely based on this world while maximizing the total reward along a certain time horizon.

\subsection{Agent Model}
The agent learns policy by exploring completely within the pre-trained environment model. Starting with a randomly selected initial state, the RL agent outputs an action, and then the state and action are fed back to the environment model to predict the next state and so on until a stop criteria. The workflow is illustrated in Fig \ref{workflow}. Unlike Atari games that usually have a clear terminal state, termination of trading actions are less well defined. Here we fix time horizon and train RL policy to maximize total reward. Once a policy has converged, we evaluate it with historical data repeatedly following similar steps: 1) from the current state with the latent LOB ($z_{t}$) and the corresponding trade prints ($u_{t}$); 2) take an action by the RL agent; 3) collect a reward; 4) if not time up, return step 1.

\section{Experiment}
We discuss details of training, testing and performance analysis in the experiment. 

\subsection{Benchmark}
We use a momentum-driven trading strategy and a classifier-based strategy as our benchmarks. The former one is a well-performing industrial strategy, and the latter one is the state-of-the-art with deep learning networks. We also use a greedy optimal strategy to measure how close the RL policy is to the optimal policy. 

\textbf{Momentum-driven Strategy} The momentum is calculated by subtracting the opening mid-price from the closing mid-price. For example, if there are 40 time ticks in one state, the closing mid-price is the mid-price at the $40^{\text{th}}$ time tick, and the opening mid-price is the one at the $1^{\text{st}}$ time tick. It roughly reflects price changes within one state and the change is assumed to be carried over to the next state. The movement can fall into three classes: up, down, and no-change w.r.t a pre-defined threshold. The trading policy is hand-crafted with a fixed quantity for each action in the action space. If the agent always takes the maximum quantity in each action either buy or sell, it is ``aggressive". The aggressive agent may have the largest reward but also has the biggest risk of losing money. The agent could also take the minimum quantity in each action. Such agent is ``conservative", has smaller risks, but it may miss significant gains.


\textbf{Classifier-based Strategy}
A classifier is trained with historical LOB data, and it predicts the mid-price movement in the next state based on the current state. Same as the momentum-based strategy, a handcrafted policy is used according to the classification, and it can be anywhere between ``progressive" and ``conservative." In this study, we build this classifier with three CNN layers and one Dense layer (64 neurons) with three output states (same configuration as the work of \cite{zhang2019deeplob}). We use the categorical cross-entropy as the loss function. The performance of the classifier is demonstrated with randomly sampled testing data in Table\ref{classifier_performance}.

\begin{table}
\centering
\caption{Classifier Performance}
\begin{tabular}{|l|l|l|l|l|}
\hline
class     & precision & recall & F1   & support \\ \hline
down      & 0.66      & 0.62   & 0.64 & 1799    \\ \hline
no change & 0.90      & 0.91   & 0.90 & 6401    \\ \hline
up        & 0.63      & 0.63   & 0.63 & 1800    \\ \hline
\end{tabular}
\label{classifier_performance}
\end{table}

\begin{figure}[h!]
\centering
\subfigure[]{\label{real_opt}\includegraphics[width=0.45\linewidth]{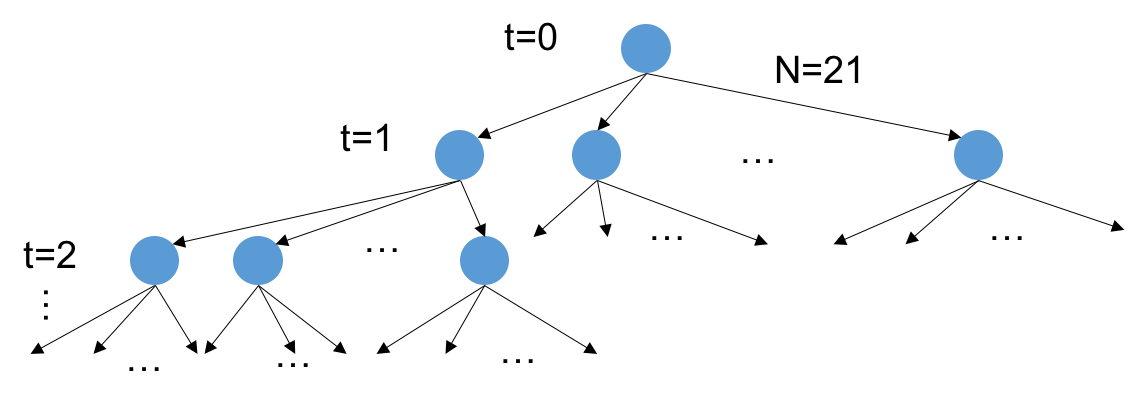}}
\subfigure[]{\label{greedy_opt}\includegraphics[width=0.45\linewidth]{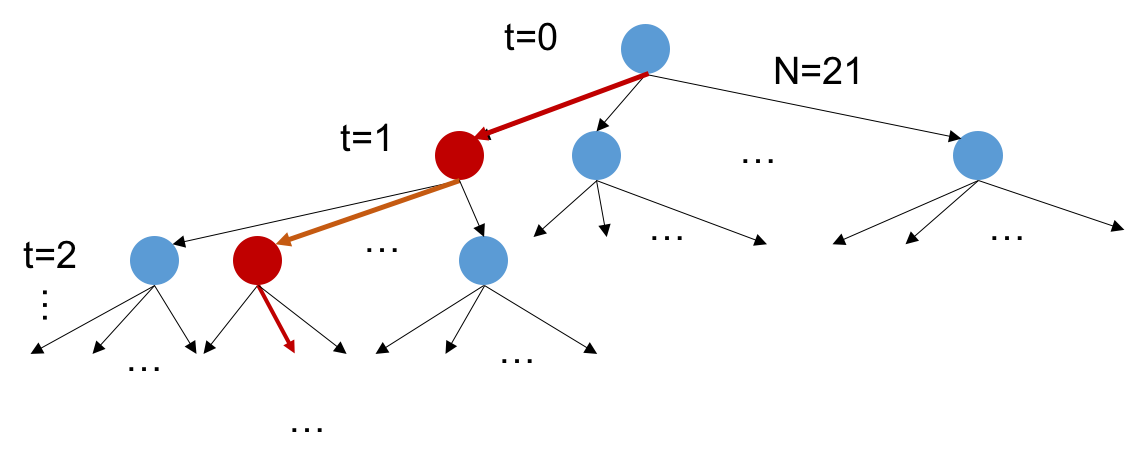}}
\caption{(a)Breadth first search for the optimal trading policy (b) greedy optimal strategy. *Nodes are states, branches are actions and each level is one time step}
\end{figure}

\textbf{Greedy Optimal}
This strategy assumes that the agent knows the future LOB in the next state. The requirement for future knowledge makes this strategy unrealistic in the real world. However, it can provide a reference for RL to measure how close the RL policy is to the optimal one. The global optimal strategy can be found with breadth-first search, demonstrated as Fig \ref{real_opt} where actions are discretized into 21 discrete actions. This strategy is computationally intractable for a trading strategy: the complexity can be above $21^{1000}$ for 1000 time ticks, and for a liquid market, a 1000-time-tick length is only $\sim$3 minutes. To overcome this obstacle, we use a greedy optimal strategy: only expand the best action with the maximum cumulative reward at each level (each time step) without visiting backward along time, shown as Fig \ref{greedy_opt}. The greedy optimal policy doesn't guarantee the global optima; however, it reduced the computational complexity from exponential to polynomial. 

\subsection{Model Architecture}

\begin{figure}[t]
 \centering
\includegraphics[width=0.3\textwidth]{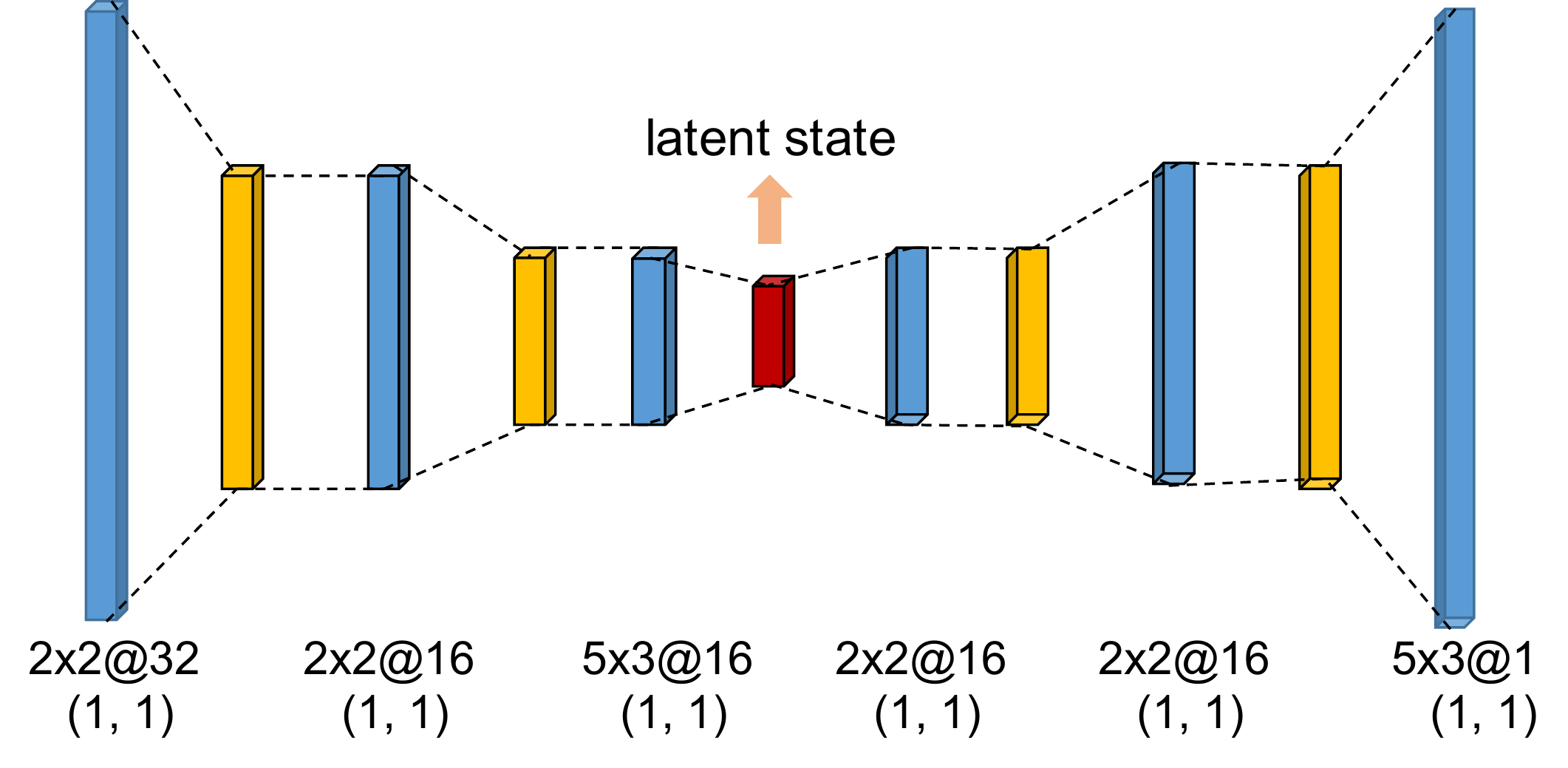}
\caption{CNN-based auto-encoder architecture.Yellow layers are Upsampling/Downsampling layers, blue layers are convolutional layers}
\label{ae_archi}
\end{figure}


We use a CNN-based auto-encoder to represent the original LOB data with latent states, and the network architecture is listed in Fig \ref{ae_archi}. The length of the trading records $u$ for each state is limited to 10 with post zero padding. In the transition model (RNN-MDN), one layer of the RNN has 128 neurons. Its input and output sequence length is $10$. We assume the state distribution is Gaussian and the number of Gaussian distributions is 5. The reward model is composed by one layer of 128 LSTM units and one layer of Dense with 40 units. The actions for greedy optimal policy and RL agent are discretized into 21 classes w.r.t the trade quantity representing $[-1000, -900, \cdots, 0, \cdots, 900, 1000]$. The training time horizon is set as 500 and each state includes 40 time ticks, and thus each training epoch covers $\sim1$h market time. The training horizon can be easily scaled up according to different demands.

\begin{figure*}[t]
\centering
\subfigure[]{\label{mid-2}\includegraphics[width=0.245\textwidth]{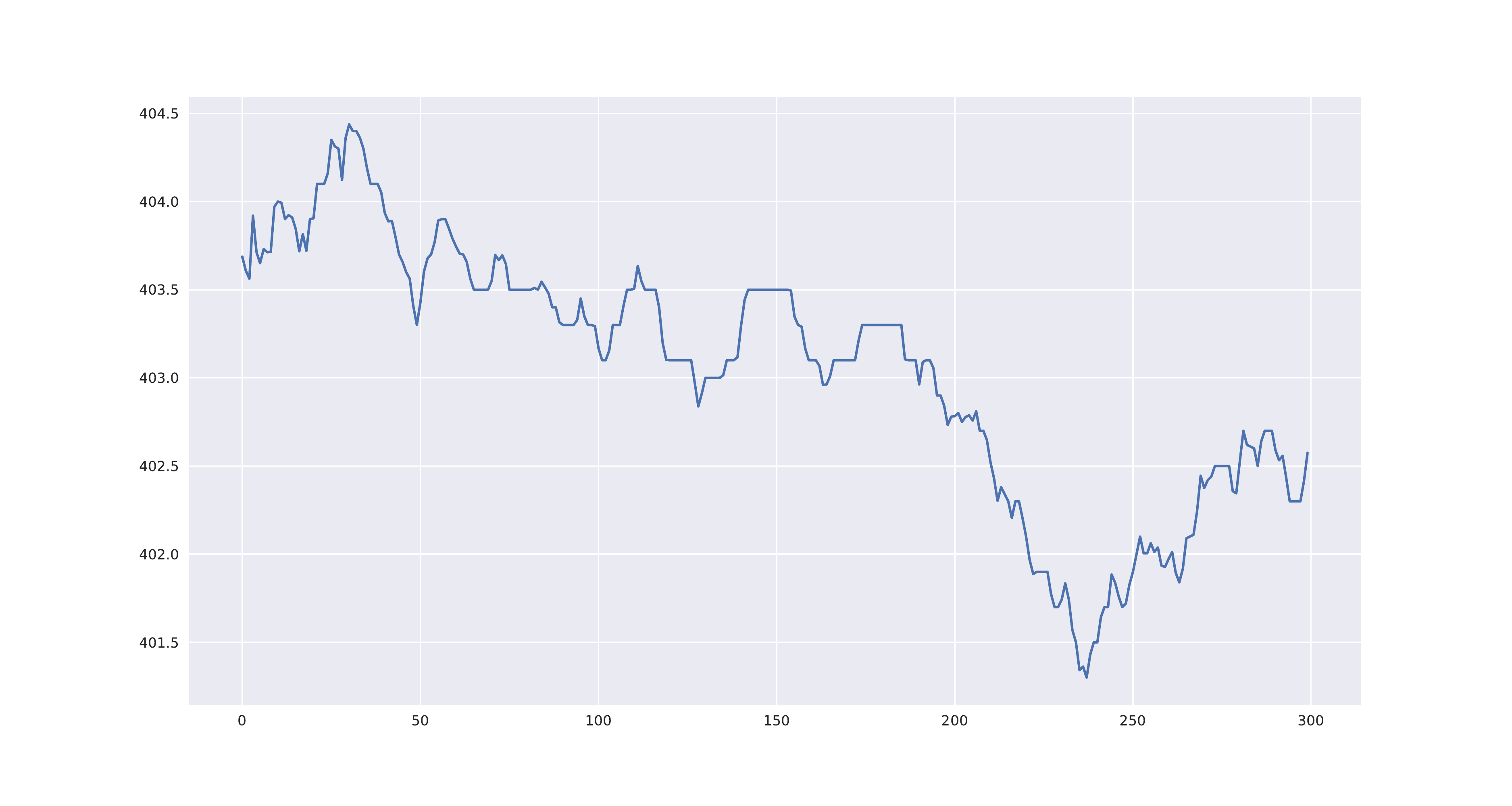}}
\subfigure[]{\label{mid-3}\includegraphics[width=0.245\textwidth]{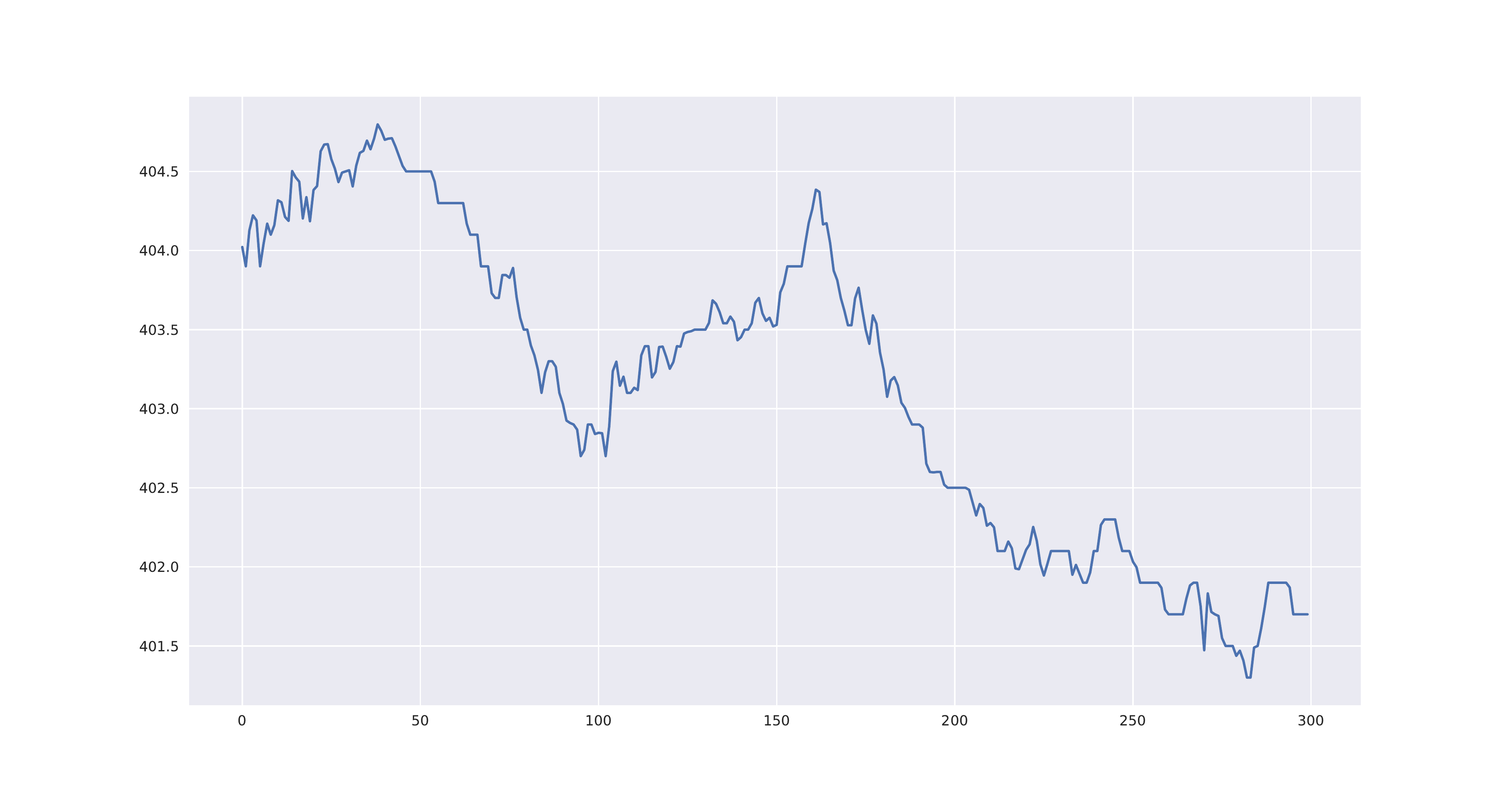}}
\subfigure[]{\label{mid-4}\includegraphics[width=0.245\textwidth]{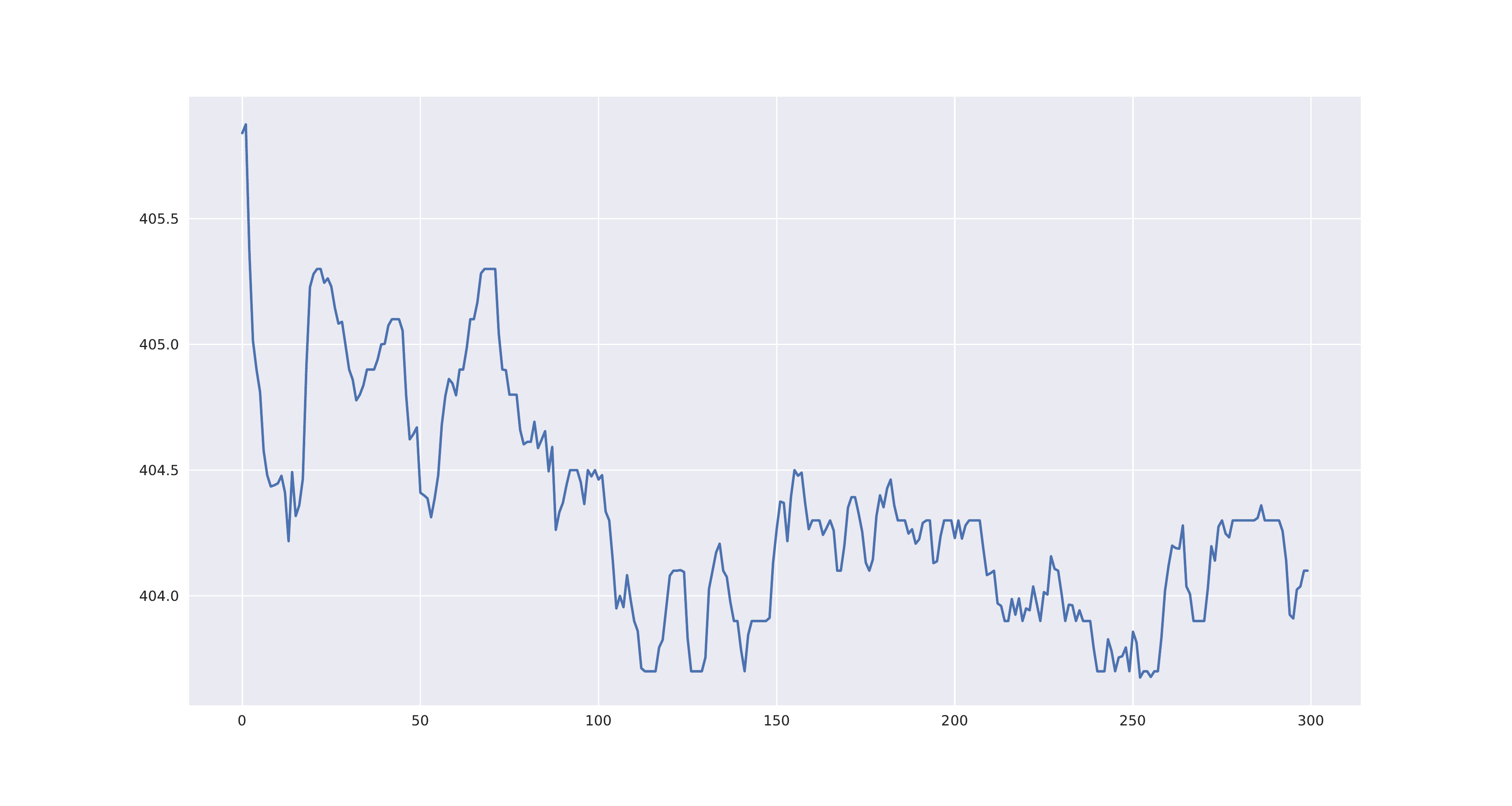}}
\subfigure[]{\label{mid-5}\includegraphics[width=0.245\textwidth]{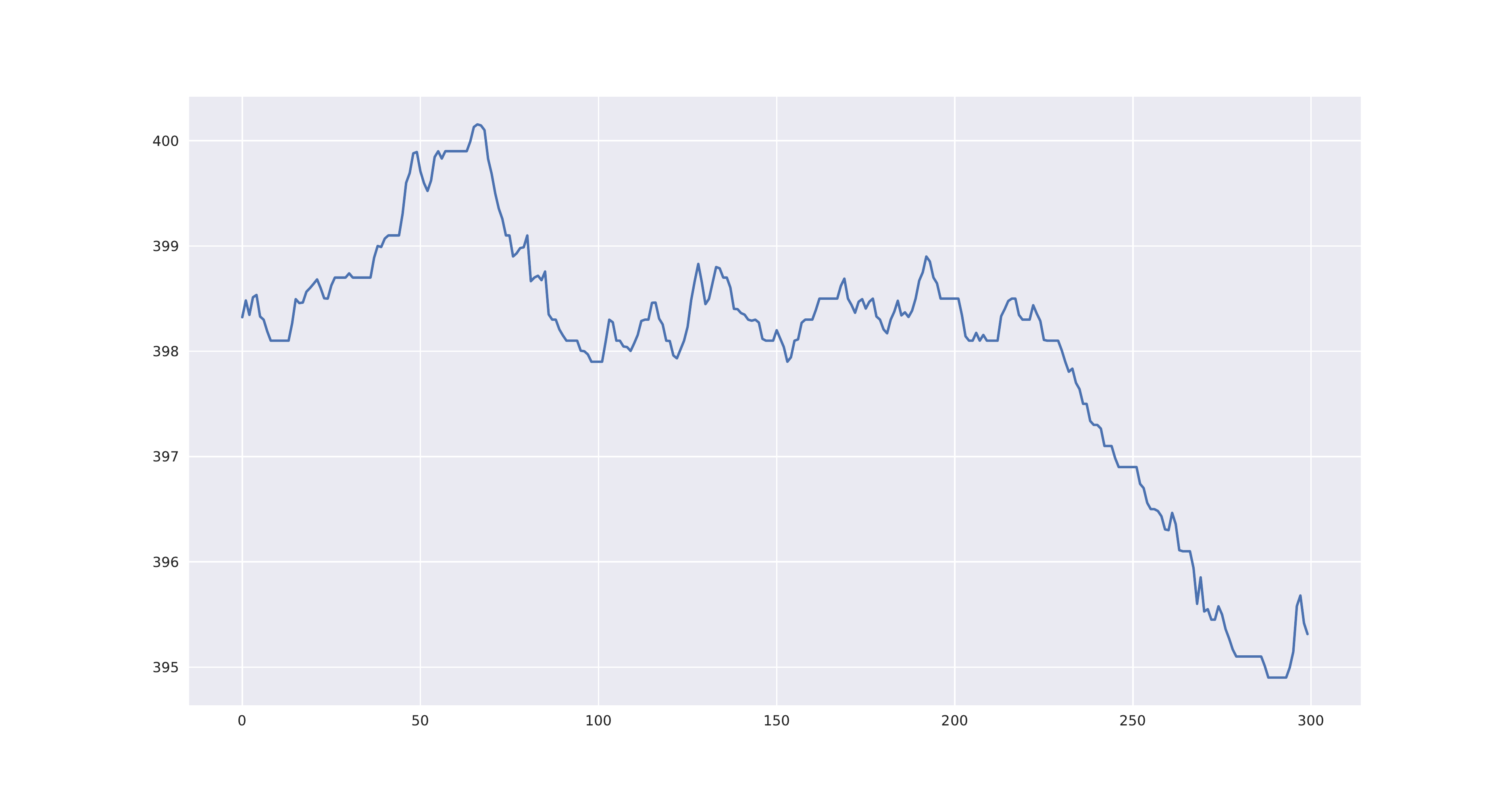}}
\\

\subfigure[]{\label{compare-1}\includegraphics[width=0.245\textwidth]{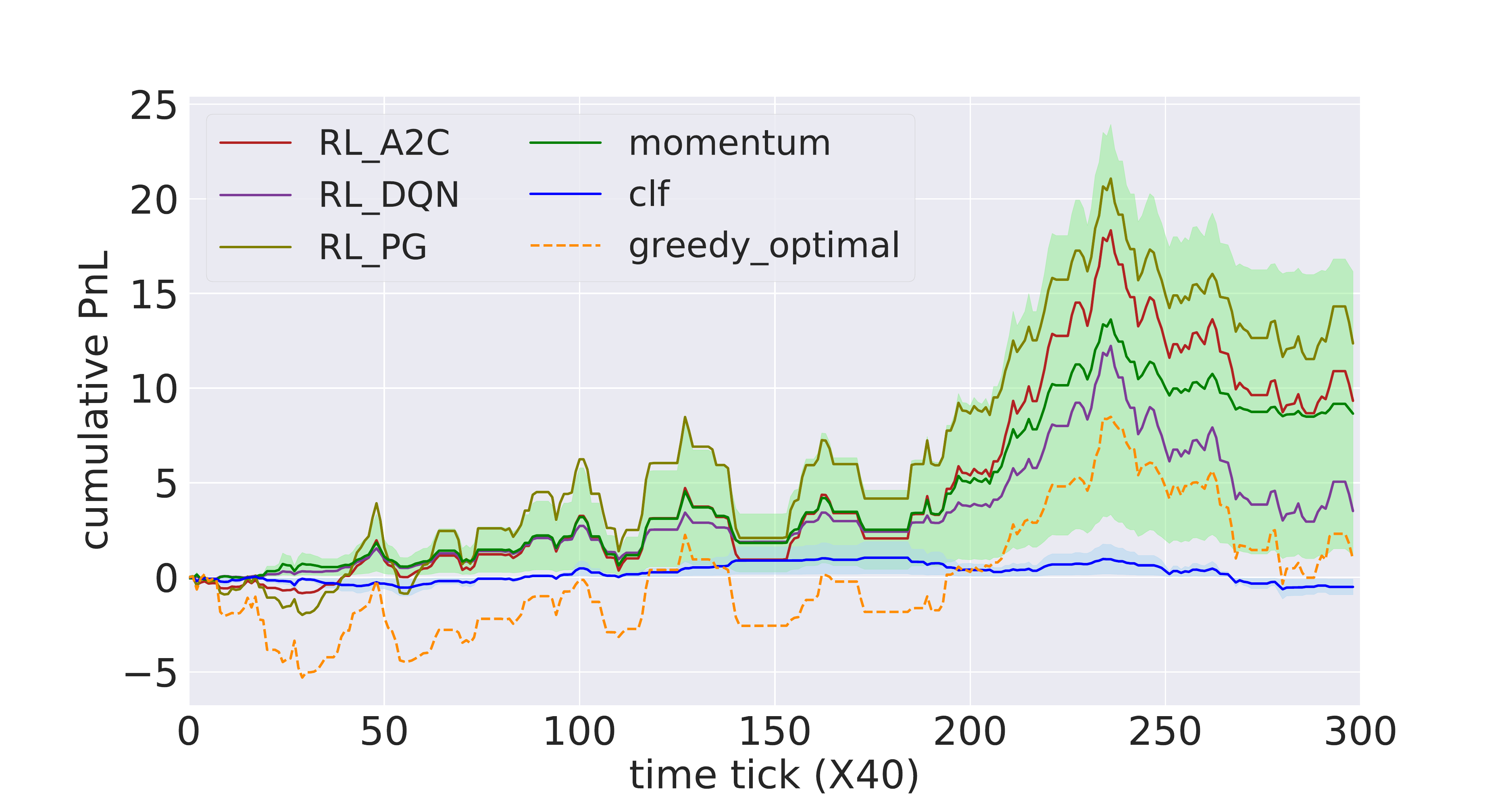}}
\subfigure[]{\label{compare-2}\includegraphics[width=0.245\textwidth]{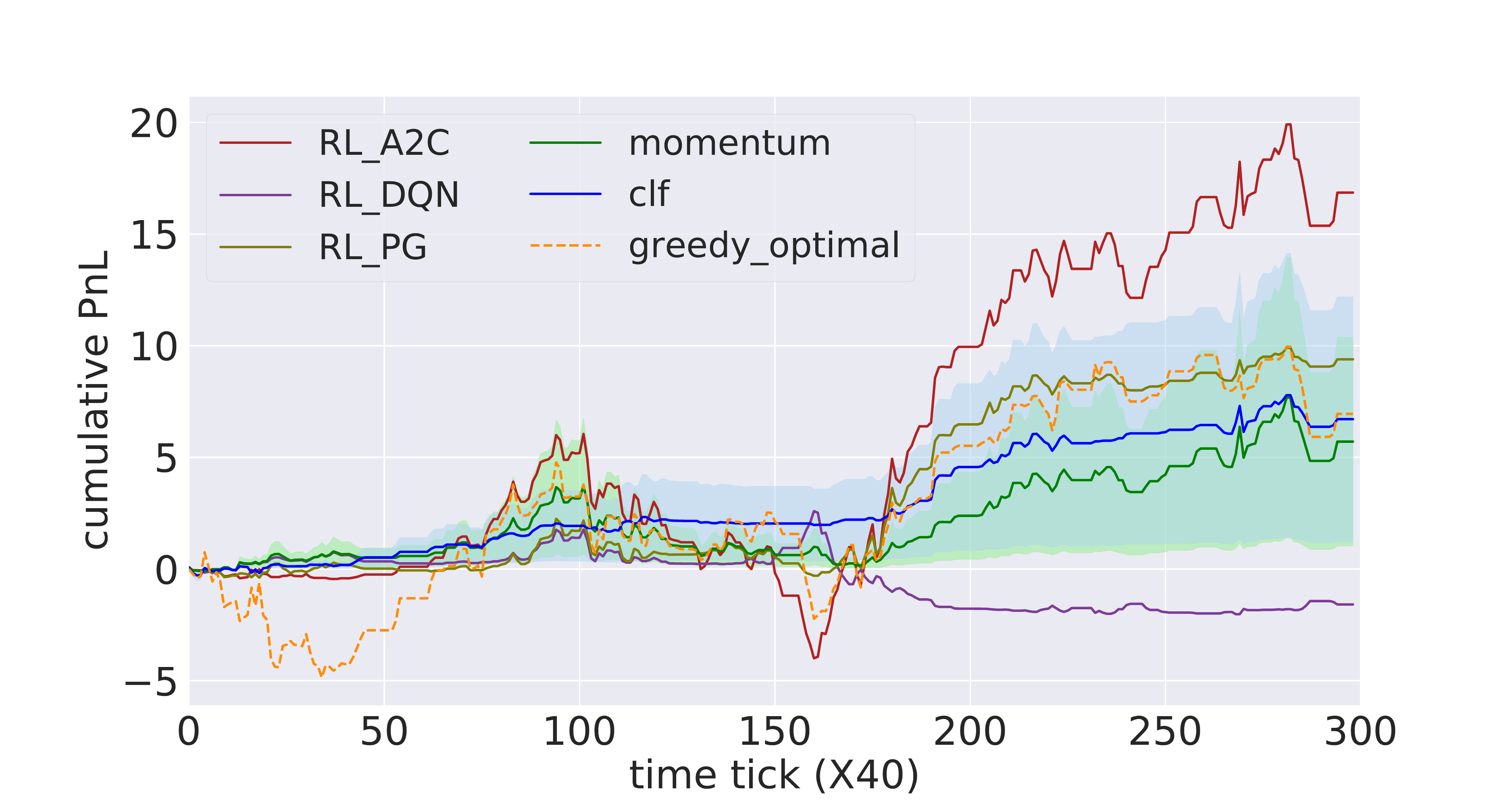}}
\subfigure[]{\label{compare-3}\includegraphics[width=0.245\textwidth]{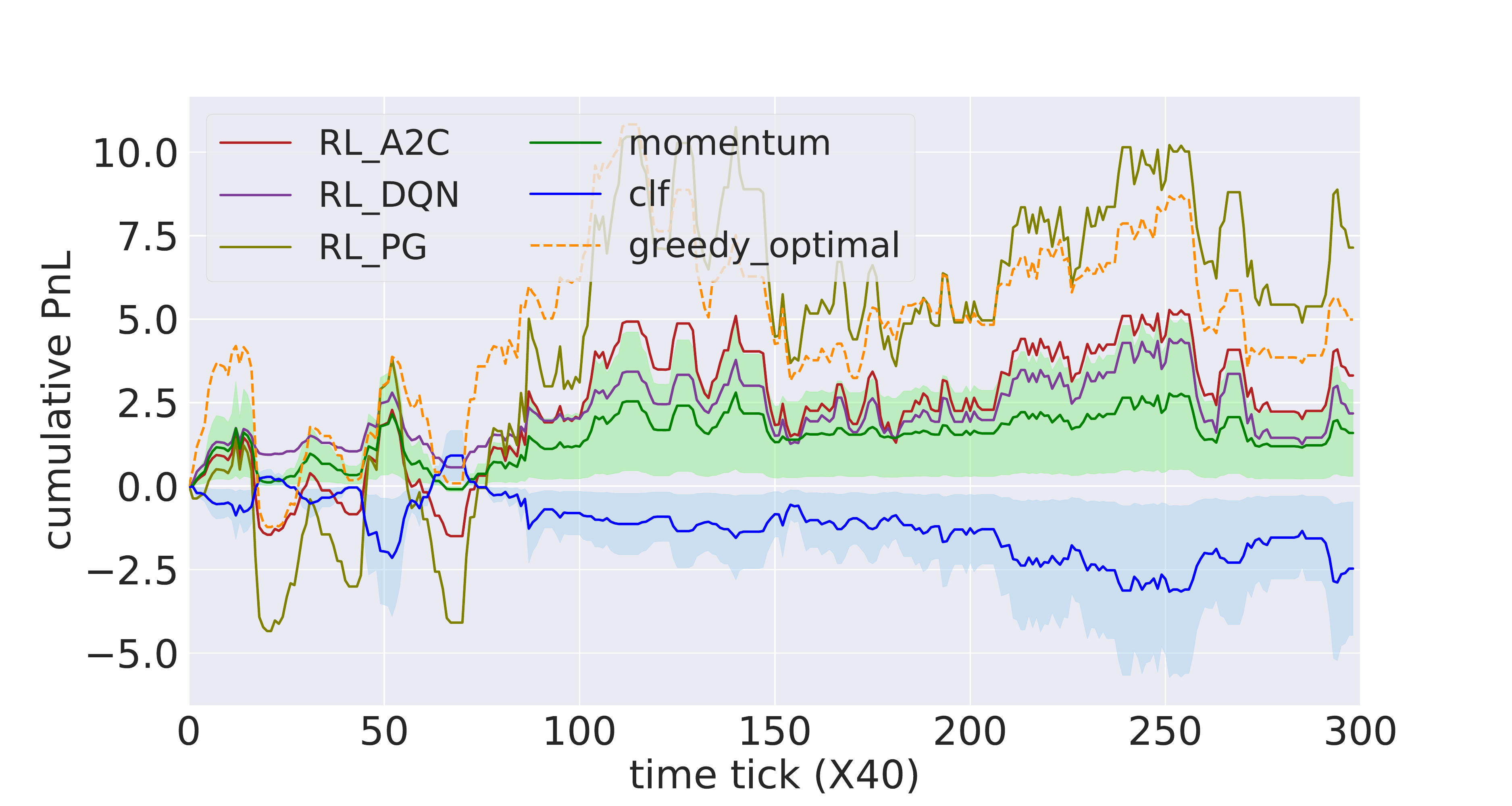}}
\subfigure[]{\label{compare-4}\includegraphics[width=0.245\textwidth]{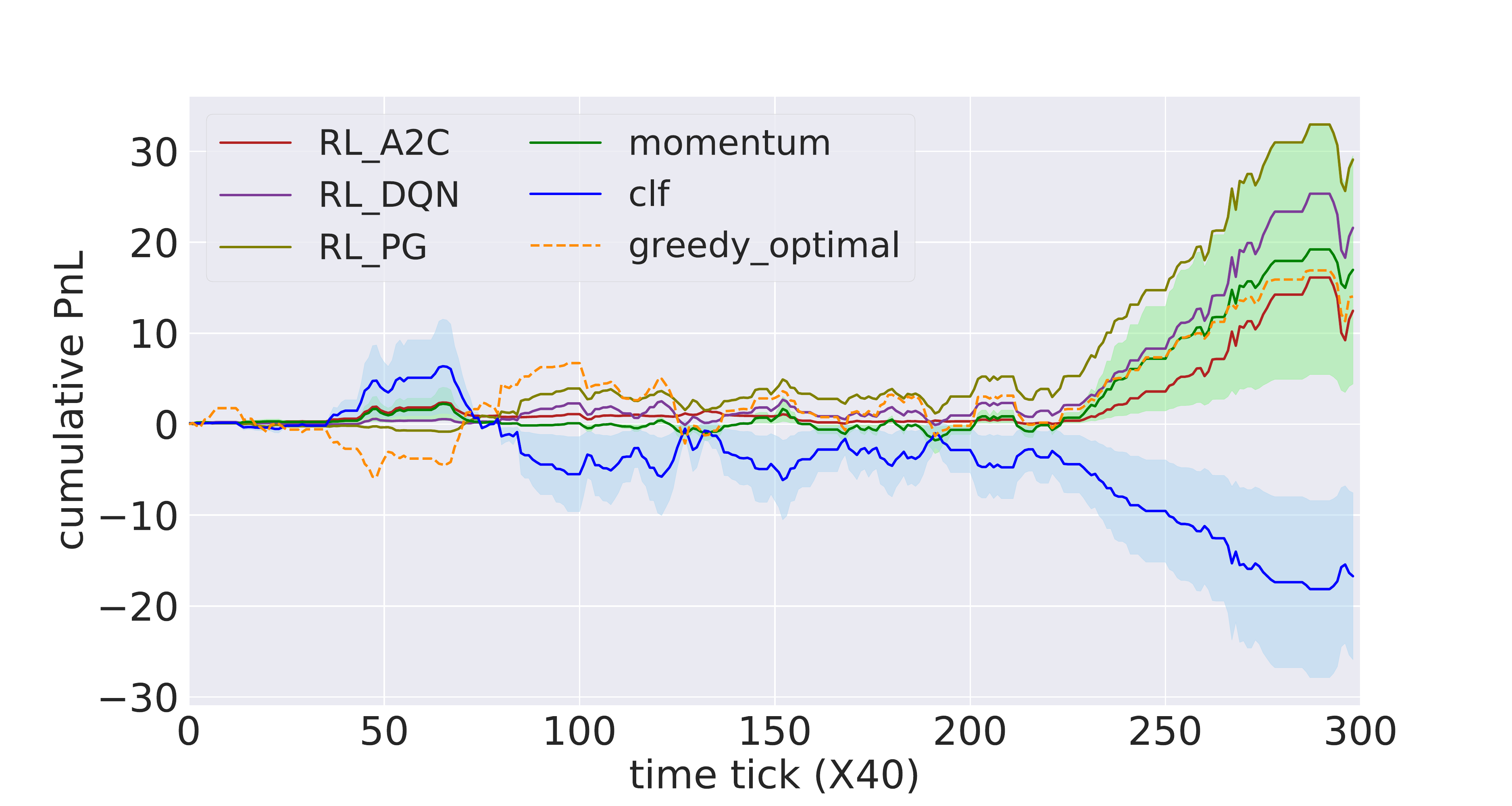}}
\caption{Trading agents performance with momentum-based agent, classification-based agent, RL-based agent and greedy optimal based agent tested on 4 random days}
\label{results_day_samples}
\end{figure*}

\begin{figure*}
\centering\subfigure[RL-based (PG)]{\label{explanation-RL}\includegraphics[width=0.24\textwidth]{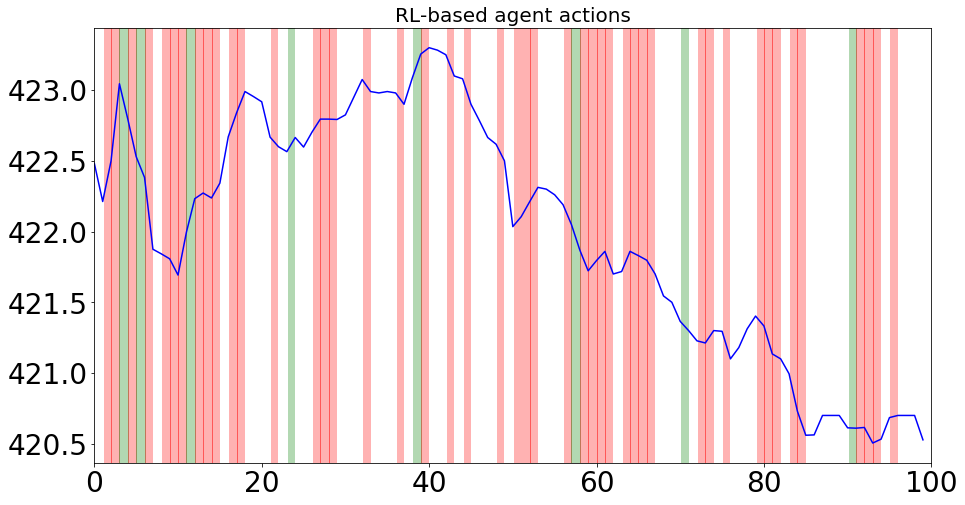}}
\subfigure[Momentum-based]{\label{explanation-momentum}\includegraphics[width=0.24\textwidth]{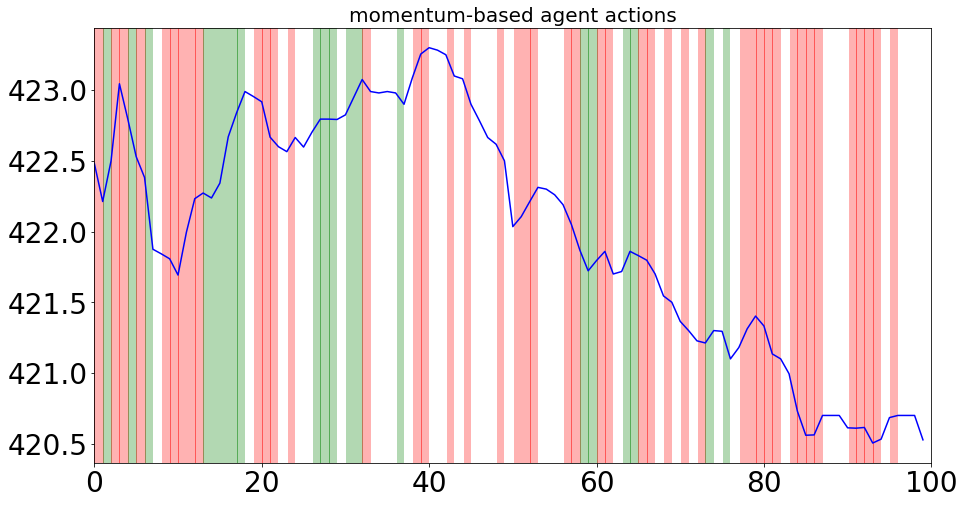}}
\subfigure[Classifier-based]{\label{explanation-clf}\includegraphics[width=0.24\textwidth]{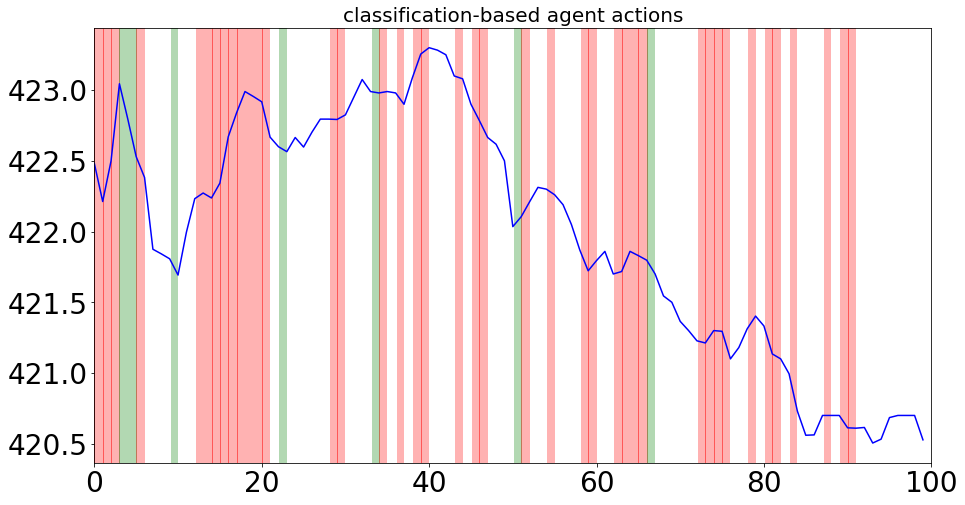}}
\subfigure[Greedy optimal]{\label{explanation-opt}\includegraphics[width=0.24\textwidth]{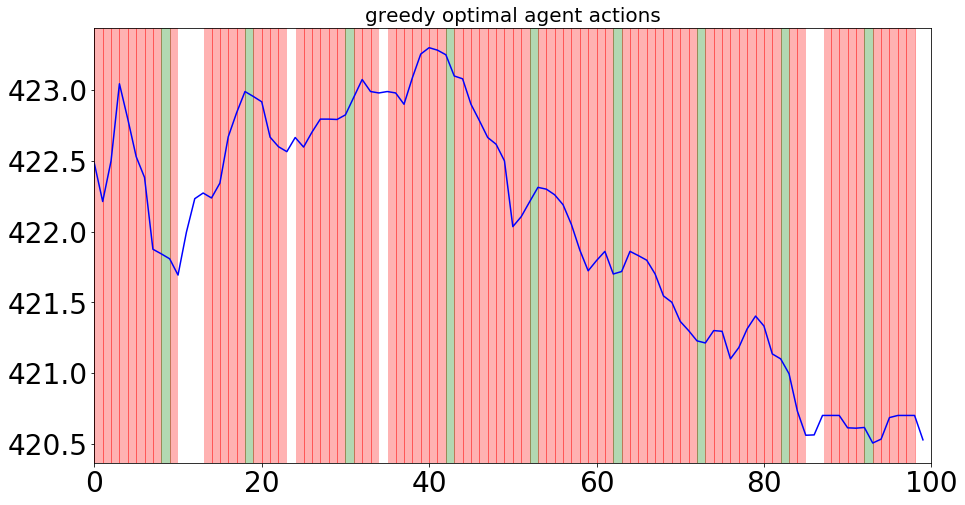}}
\caption{Performance explanation in terms of actions selection: red shadows represent ``sell", green shadows represent ``buy", no actions taken in blank area. The blue line is the corresponding mid price.}
\label{action selection}
\end{figure*}

\subsection{Experiment Results}


\begin{figure}[h!]
\centering
\subfigure[]{
\includegraphics[width=0.45\linewidth]{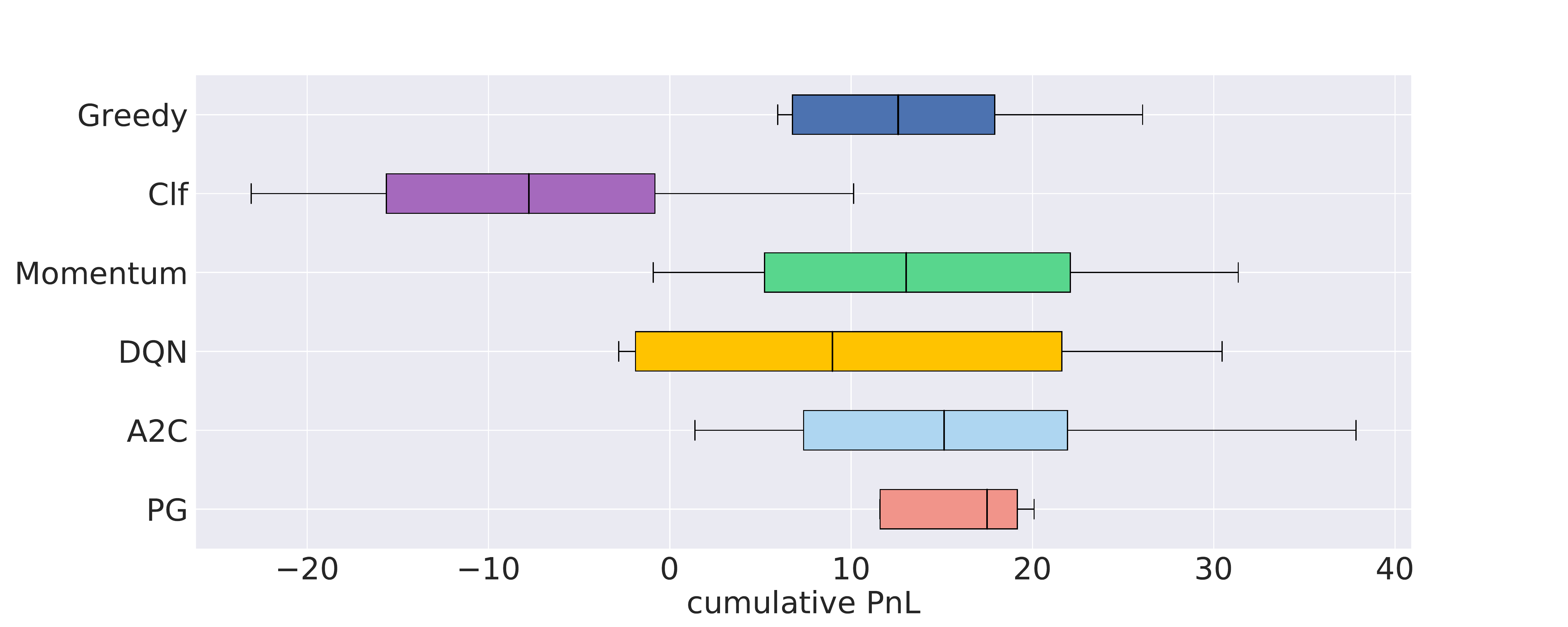}
\label{var-descending}}
\subfigure[]{
\includegraphics[width=0.45\linewidth]{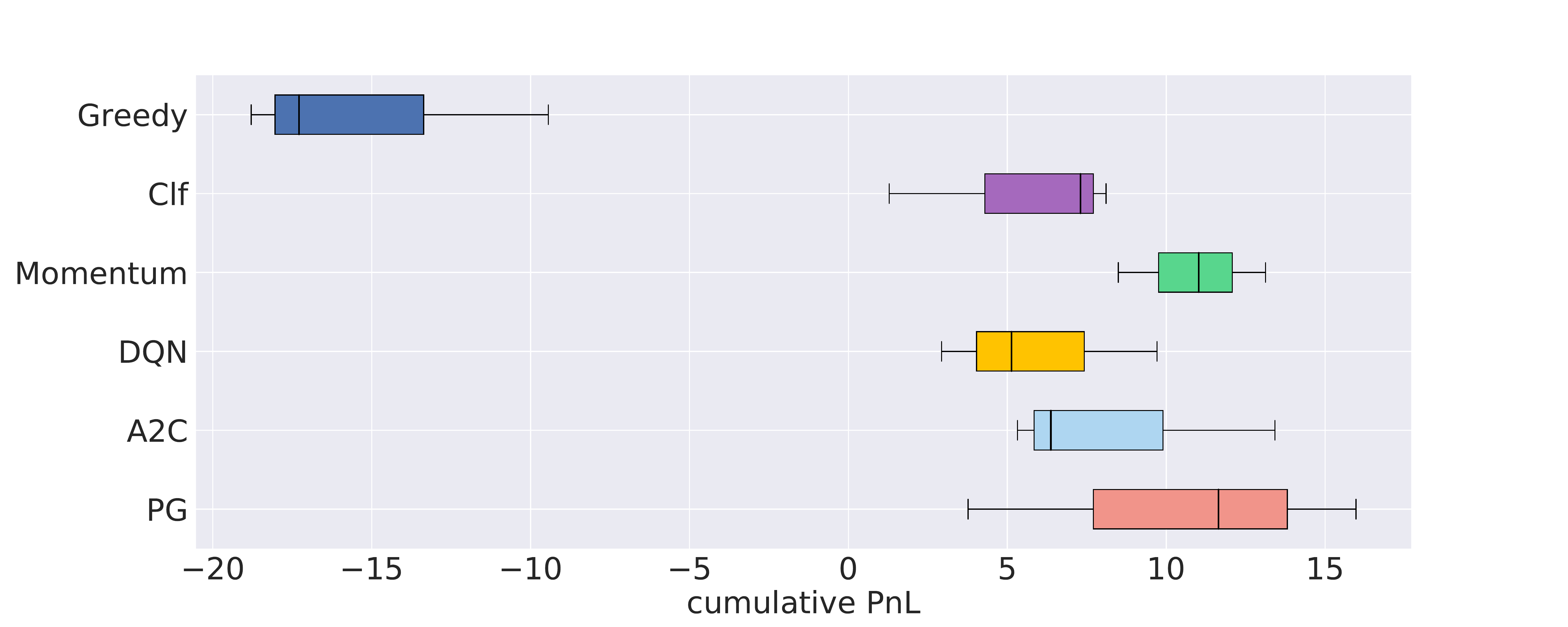}
\label{var-ascending}}
\subfigure[]{
\includegraphics[width=0.45\linewidth]{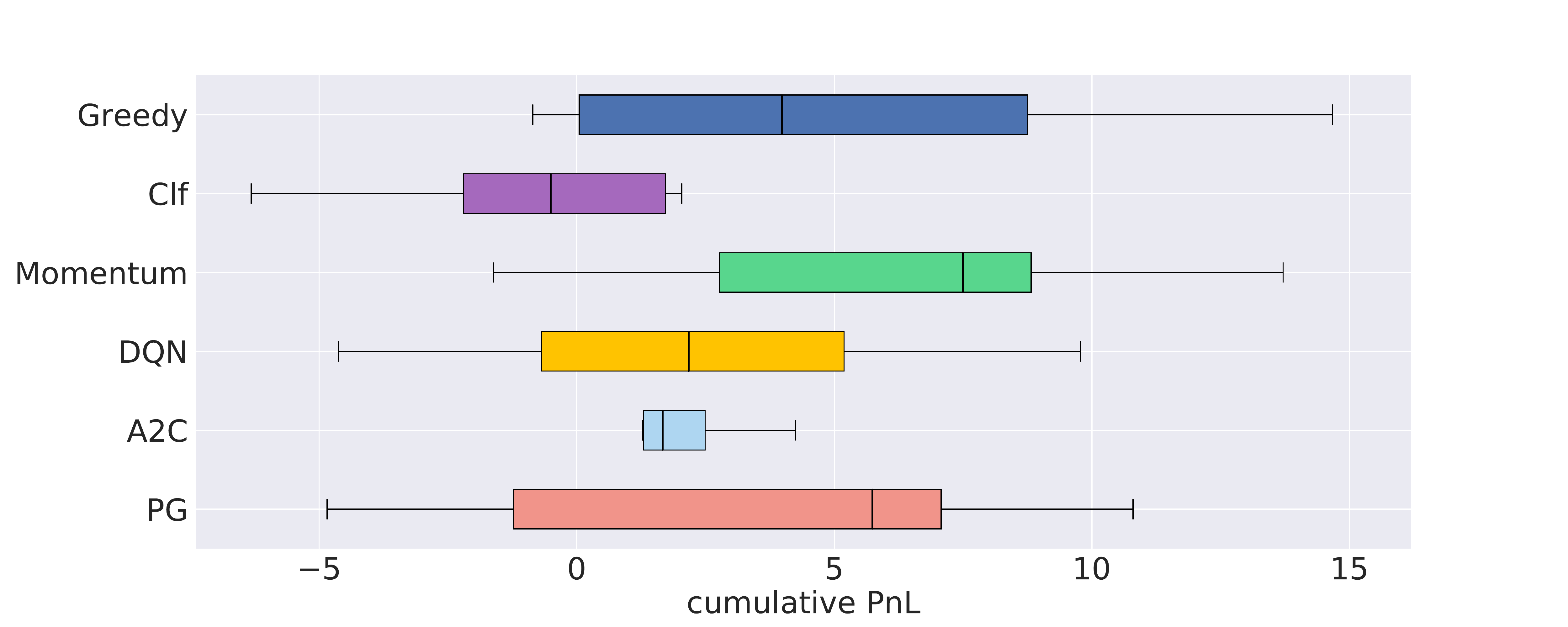}
\label{var-osci}}
\subfigure[]{
\includegraphics[width=0.45\linewidth]{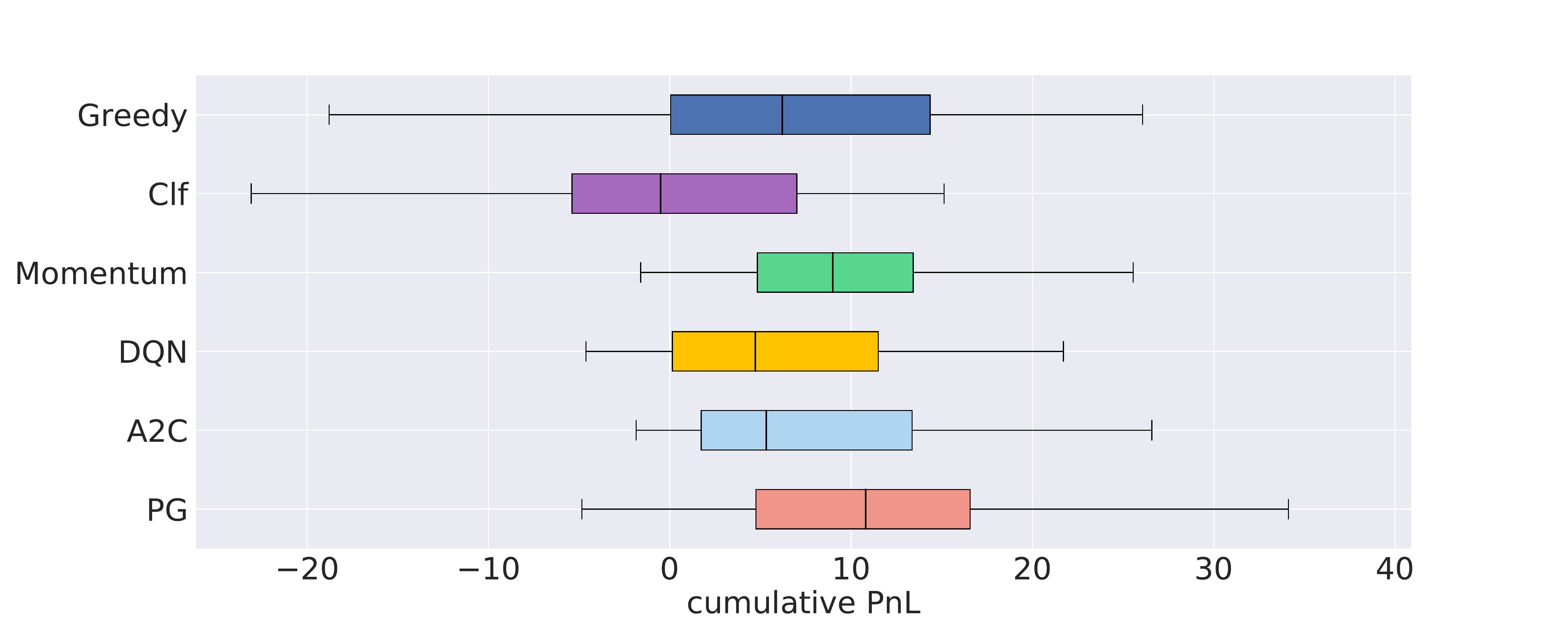}
\label{var-total}}
\caption{performance variance in the descending market as \ref{var-descending}, ascending market as \ref{var-ascending}, oscillating market as \ref{var-osci}, and total variance with varied market trending as \ref{var-total}.
}
\label{performance_variance}
\end{figure}

We randomly picked four days in April 2018 with $300 \times 40$ time ticks ($\sim 0.56$hr) each day to demonstrate with the corresponding mid-price shown on top, seen in Fig \ref{results_day_samples}. We considered the transaction fee as 2\% of PnL per quantity unit. The shadow with the momentum-based and the classifier-based strategy represents all possible performances between progressive and conservative policies, and its upper edge shows the best performance at any given time tick, however, the policy may not always be the same. For the sake of reducing the risk and increasing the cumulative rewards, a mixed strategy is more reasonable. However, it requires more complex hand-crafted policies with benchmarks. Alternatively, the RL policy is a straightforward approach to a mixed strategy where actions with various quantities are taken at different states. 

As shown in Fig \ref{results_day_samples}, the RL agent's performance is very close to the greedy-optimal solution in general and outperforms the other two benchmarks' average performance by $\sim$ 10\% to 30\%. In some time segments, the progressive momentum-based agent beats our RL agent (e.g, $t=100 \sim t=200$ in fig \ref{compare-2}. This is because the quantity in each RL action during this period is smaller than the momentum-based agent to avoid the future risk predicted by the transition model and that also explains why the cumulative PnL surpasses the momentum-based after $t=200$ when the mid-price starts dropping. A2C overall provides more stable performance compared to PG and DQN. DQN sometimes performance poorly, and this can be improved by having a better state representation (will be addressed in our future work). The classification-based agent performs relatively poorly, and we believe it is caused by the classification bias. The classifier tends to misclassify the movement as ``no change" because this is the dominating class, and thus the classifier-based agent takes fewer actions. This is actually a real-world bottleneck for classification methods in finance because the market doesn't have a big change within a short period in general. One of the improvement solutions is to extend the window length of each state so that each state expands a longer time horizon, and more changes may be captured. However, it will decrease the action granularity because only one action is associated with one state. 

\subsection{Explanation and Analysis}

To understand and explain the performance better, we also visualize the action selections in Fig \ref{action selection} with mid-price highlighted in blue. For observation convenience, we only showed the first 100 time ticks. Three agents (RL(PG), momentum-based, and classifier-based) have similar action frequency. The RL agent's strategy can be summarized as always sell around the local peaks and buy around the local valleys. The momentum-based policy has higher action frequency to switch the ``buy" and  ``sell" actions, and this may cause lower position so that smaller rewards are collected sometimes. The classifier-based agent has relatively lower action frequency, and it takes irrational actions sometimes, such as ``buy" at the local peaks in the first five time ticks. The greedy optimal has the highest trading frequency, and its main strategy is to maintain a high position and not to take actions when the one-step PnL is negative. We also compared the policy performance variance with different methods in descending markets (Fig \ref{var-descending}), ascending market (Fig \ref{var-ascending}), oscillating market (Fig \ref{var-osci}) and overall (Fig \ref{var-total}). Ascending market means mainly goes up along time (and similarly for the others). RL performs very well in descending and ascending markets, much better compared to the start-of-the-art classifier-based approach. However, it doesn't perform well in the oscillating market,  caused by insufficient training data (only $\sim$15.8\% in total training data). We'll address this problem in our future work.

\begin{figure}
    \centering
    \includegraphics[width=0.25\textwidth]{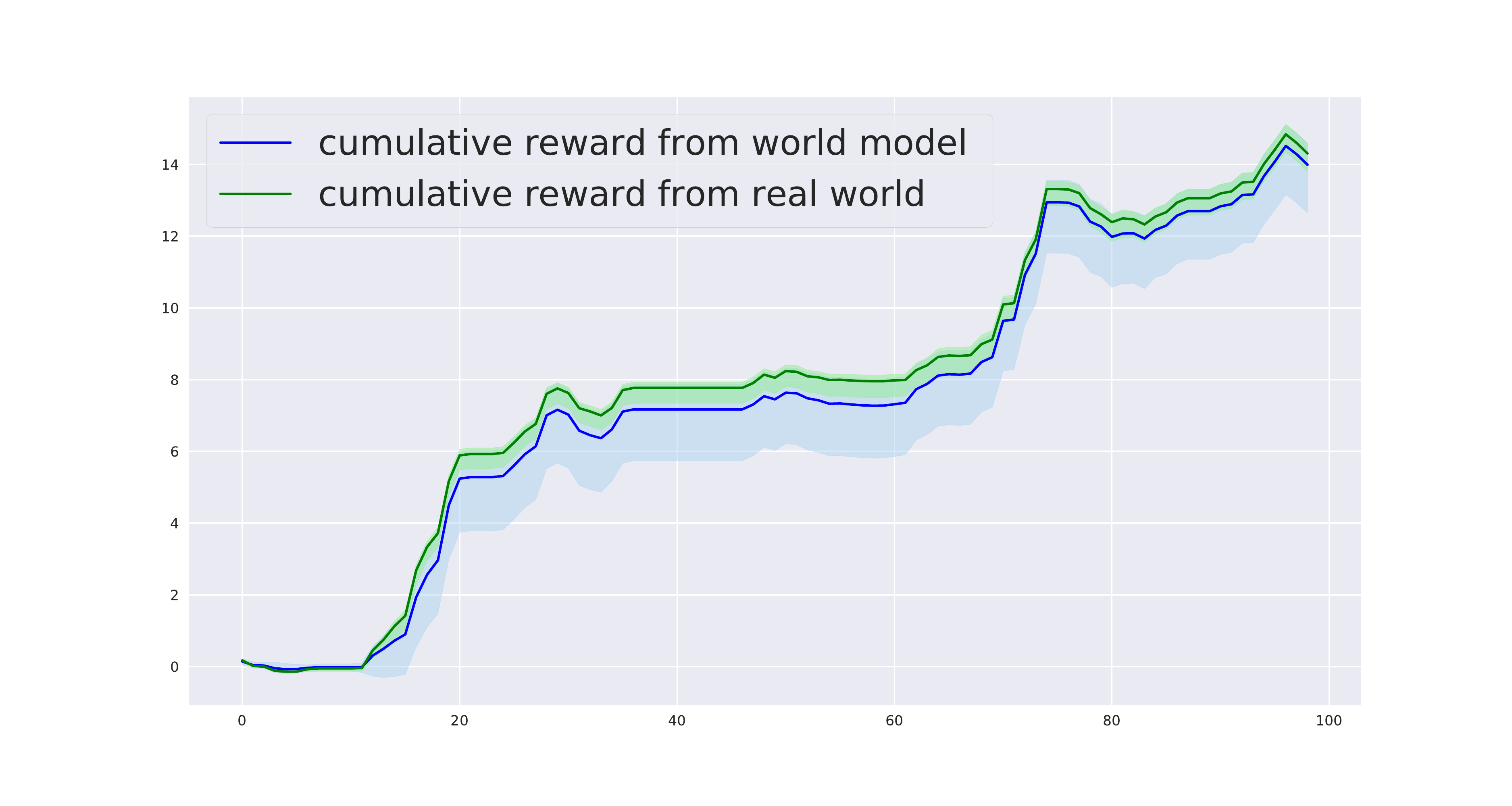}
    \caption{RL policy (PG) transferability: the policy trained fully based on the environment model has acceptable performance in the real world }
    \label{transfer}
\end{figure}

We also show the transferability of RL policy from the world model to the real environment. With five randomly picked days, the same policy is implemented in both world model and real environment on each day, then we compare the cumulative reward. As shown in Fig \ref{transfer}, the asymptotic performance (cumulative reward) of RL policy shows the transferability of RL policy from the environment model to the real world. 

\section{Conclusion and Future Work}
We propose a model-based RL for learning a trading policy in finance. Our work provides the potential of RL applied in the domains where the state space is high dimensional, and real environment interactions are expensive or infeasible. We also contribute a framework for modeling trading markets for future purposes. The trading market is a complex domain, and more dynamic factors should be considered, including broker fees, dynamic transition fees,  etc. Some hand-crafted rules could be combined with an RL agent (e.g., setting a risk threshold). System time latency is another concern: a delayed response may influence trading policy's efficacy. Different market data with different liquidity  should be tested with our RL approach to demonstrate trading strategy robustness. 


\bibliography{ref} 
\bibliographystyle{aaai}
\end{document}